\documentclass[conference]{IEEEtran}
\IEEEoverridecommandlockouts
\usepackage{cite}
\usepackage{amsmath,amssymb,amsfonts}
\usepackage{algorithmic}
\usepackage{graphicx}
\usepackage{textcomp}
\usepackage{comment}
\usepackage{xcolor}
\usepackage{lettrine}
\usepackage{hyperref}
\usepackage[T1]{fontenc}
\def\BibTeX{{\rm B\kern-.05em{\sc i\kern-.025em b}\kern-.08em
    T\kern-.1667em\lower.7ex\hbox{E}\kern-.125emX}}
\begin{document}

\title{3CSim: CARLA Corner Case Simulation for Control Assessment in Autonomous Driving}
\author{
\IEEEauthorblockN{Mat\'{u}\v{s} Čávojský\textsuperscript{1}, Eugen \v{S}lapak\textsuperscript{2}, Mat\'{u}\v{s} Dopiriak\textsuperscript{2}, Gabriel Bugár\textsuperscript{1}, Juraj Gazda\textsuperscript{2}}
\IEEEauthorblockA{\textsuperscript{1}Department of Electronics and Multimedia Telecommunications, Technical University of Ko\v{s}ice, Slovakia}
\IEEEauthorblockA{\textsuperscript{2}Department of Computers and Informatics, Technical University of Ko\v{s}ice, Slovakia}
Email: \{matus.cavojsky, eugen.slapak, matus.dopiriak, gabriel.bugar, juraj.gazda\}@tuke.sk
}
\maketitle

\begin{abstract}
We present the CARLA corner case simulation (3CSim) for evaluating autonomous driving (AD) systems within the CARLA simulator. This framework is designed to address the limitations of traditional AD model training by focusing on non-standard, rare, and cognitively challenging scenarios. These corner cases are crucial for ensuring vehicle safety and reliability, as they test advanced control capabilities under unusual conditions. Our approach introduces a taxonomy of corner cases categorized into state anomalies, behavior anomalies, and evidence-based anomalies. We implement 32 unique corner cases with adjustable parameters, including 9 predefined weather conditions, timing, and traffic density. The framework enables repeatable and modifiable scenario evaluations, facilitating the creation of a comprehensive dataset for further analysis.
\end{abstract}

\begin{IEEEkeywords}
autonomous driving, CARLA simulator, corner cases, reinforcement learning.
\end{IEEEkeywords}

\section{Introduction}
\label{Introduction}
\lettrine{A}{utonomous} driving (AD) has made substantial progress in recent years, driven by advancements in computational hardware, the availability of high-quality data, and the development of sophisticated sensors such as cameras and LiDARs. The integration of deep learning (DL) techniques has further accelerated the evolution of AD systems. Notable achievements include advancements in perception tasks \cite{chib2024recent} serving as a foundation for other modules for planning \cite{luo2022jfp, siu2023motion} and control through deep reinforcement learning (DRL) \cite{ravi2022deep, jingwei2023eventtriggered} or imitation learning \cite{jianyu2019deep, siyu2023hierarchical}. These technological innovations have positioned major companies, such as Tesla and Waymo, to bring autonomous vehicles (AVs) into real-world traffic. However, Waymo autonomous taxis have demonstrated unusual behavior, such as stopping when a pedestrian wears a T-shirt with a printed STOP sign. This raises concerns about the system's ability to generalize effectively, as well as its safety and reliability. \par
AD systems perform well in routine situations commonly represented in their training data. However, the real world presents a wide range of rare and unpredictable scenarios, known as corner cases, which are not typically encountered during training. These corner cases, such as a pedestrian suddenly emerging from behind an obstruction or unexpected road obstacles, deviate significantly from standard traffic conditions \cite{bogdoll2022anomaly, hanh2023adslead}. As a result, AVs often struggle to generalize effectively in these situations, leading to potential failures in perception and control. While advancements in artificial intelligence (AI) have improved AD systems, the limited representation of corner cases in training datasets remains a critical challenge \cite{bogdoll2021description, fu2024drive}. Addressing this issue requires comprehensive dataset engineering and the inclusion of diverse corner case scenarios to enhance system robustness and safety. However, obtaining relevant real-world data for these rare and often dangerous scenarios is exceptionally difficult.\par
Simulators are crucial in the development of AD systems, offering controlled environments for safe and efficient training and testing. The CARLA simulator, widely used for AD research \cite{hossain2023autonomous, nehme2023safe, kashyap2023transfuser}, allows for the generation of necessary data without the risks of real-world scenarios. These environments equip AD systems to manage routine scenarios. However, the effective handling of rare and unpredictable events in real-world contexts remains a significant challenge yet to be fully implemented. By allowing the repetition of scenarios with varied conditions, simulators like CARLA facilitate comprehensive evaluation and better prepare AD systems for real-world deployment, despite the greater complexity of actual driving conditions. \par
This paper introduces the CARLA corner case simulation (3CSim) for evaluating AD systems in a controlled and deterministic environment. The simulation allows for the repetition of identical or slightly varied events, creating unique scenarios to compare the behavior of advanced AD systems, which is not feasible in real-world settings. Designed to assess the control capabilities of AD systems in rare and unpredictable traffic situations, the 3CSim also enables systematic data collection from these events to build a dataset for further analysis. Additionally, we propose a taxonomy to classify the implemented corner cases into state anomalies, behavior anomalies, and evidence-based anomalies. \par

The main contributions
of this paper are summarized as follows:
\begin{enumerate}
    \item We propose 3CSim for control assessment of AD models in controlled and deterministic environment.
    \item We propose a taxonomy of advanced corner case scenarios implemented in our proposed simulation. 
\end{enumerate}

\section{Related Work}
\label{related_work}

In this section, we compare our approach to similar methods aimed at simulating corner case scenarios for enhancing generalization of AD systems. Additionally, we evaluate our taxonomy and terminology against other proposed classifications and terms.

\subsection{Synthesizing Corner Cases} 

This subsection examines methods for synthesizing corner cases with simulation tools, relevant to our approach of generating unique scenarios for AD model evaluation. \par

Research \cite{niu2023stackelberg} presents the Stackelberg driver model (SDM), which uses a scenario-based framework to generate safety-critical corner cases. It models AV-other vehicle interactions as a Stackelberg game, allowing iterative policy refinement in response to challenging behaviors. The paper \cite{drayson2024ccsgg} uses heterogeneous graph neural networks (HGNNs) to generate corner cases by perturbing scene graphs in driving simulators like CARLA, achieving 89.9\% prediction accuracy. This method effectively tests AV robustness and enhances safety validation. In contrast, our approach involves manually generating corner cases with varied input conditions to ensure diversity of events. \par

CornerSim \cite{daoud2024cornersim} is a framework for generating synthetic corner-case scenarios to test AD systems. It facilitates the creation and modification of diverse driving scenarios, producing raw sensor data and labeled datasets for training and validation. By simulating rare and challenging situations, CornerSim enhances the robustness of perception systems in AVs and provides data, such as the CornerSet dataset, for benchmarking detection algorithms and improving testing and development. Unlike our approach, which focuses on control, CornerSim's data is primarily for perception modules.\par

The paper \cite{li2024first} introduces a first-principles sensor model integrated into the CARLA simulator to enhance safety of intended functionality (SOTIF) testing for AVs particularly in adverse weather conditions like fog. Additionally, it presents a meta-heuristic algorithm to efficiently identify corner cases by reducing the scenario search space. This method improves corner case detection and resolves synchronization issues between simulators and autonomous systems. However, it provides only limited set of representative corner cases in contrast to our approach.

\subsection{Corner Case Taxonomies}
Corner case taxonomies can be divided into two main groups. The first group focuses on neural network interpretation, as described in \cite{zhou2023corner}. The second group is concerned with visual applications. Breitenstein et al. \cite{breitenstein2020systematization} systematize corner cases by abstracting events from the pixel level to complex scenarios, a framework widely adopted by other taxonomies. Bogdoll et al. \cite{bogdoll2021description} define corner cases as critical events that are infrequent during AI model training, while Heidecker et al. \cite{heidecker2021applicationdriven} extend this definition to include data from cameras, LiDAR, and radar. Pfeil et al. \cite{pfeil2022on} categorize corner cases based on environmental factors, functional constraints, and system-internal conditions. Although our taxonomy is also based on visual applications, it specifically focuses on the classification of corner cases by their unusual states, behaviors, or pre-event indicators, as detailed in section \ref{sec:taxonomy}. \par

\section{CARLA Corner Case Simulation}
\label{c3_framework}
\begin{figure}[!ht]
	\centering
\includegraphics[width=0.48\textwidth]{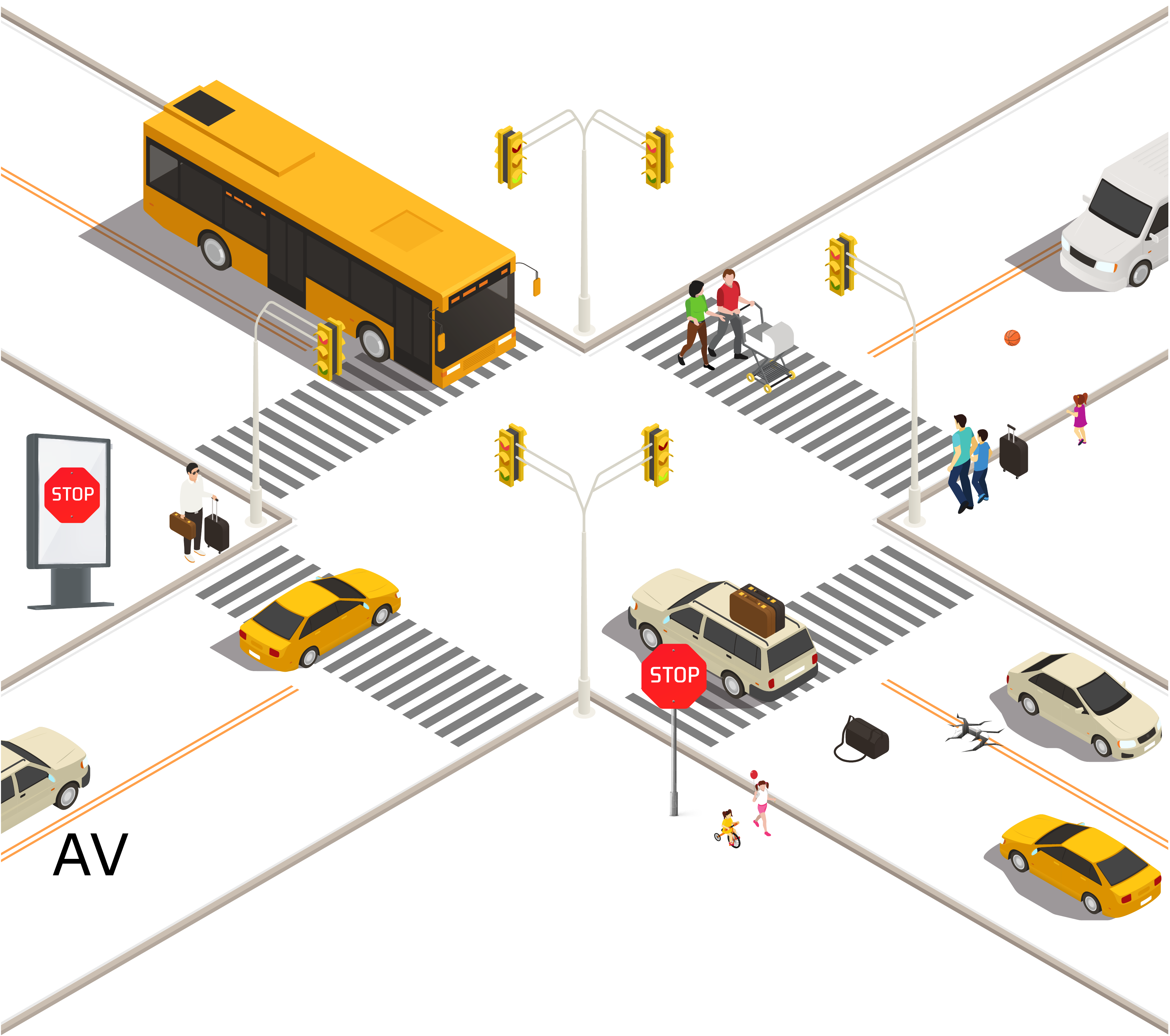}
    	\caption{A wide range of corner cases that can arise in real-world traffic scenarios.}
	\label{fig:corner_cases}
\end{figure}

The deployment of AVs in real traffic environments demands a thorough understanding and effective handling of various driving scenarios, including corner cases. As depicted in Fig. \ref{fig:corner_cases}, advertisements obscuring STOP signs can lead to incorrect decisions by AV. A ball on the road may signal the presence of children who could run onto the street without regard for safety. An unsecured suitcase on a vehicle may fall, requiring prompt reaction by AV. Additionally, AV often faces complex situations, such as navigating between a pothole on the right and children on the left, necessitating the optimal choice of action. The behavior of AVs in such scenarios, along with their potential weaknesses, remains a concern. Even minor variations in traffic situations can overwhelm the AD system, significantly undermining its ability to achieve high reliability.\par

We introduce the 3CSim, specifically designed to simulate a wide range of corner cases for AD systems in urban environments. This simulation facilitates deterministic simulations, allowing for the precise repetition of complex events under identical conditions. It also permits the controlled variation of factors such as weather, vehicle speed, and pedestrian behavior to assess the performance of advanced AD systems. These systems, modeled as AVs, are integrated into the simulation to evaluate their ability to generalize effectively to rare and challenging scenarios, known as corner cases. The 3CSim can also generate extensive dataset by running simulations without any AD systems, providing valuable insights for analyzing these corner cases. By enabling comprehensive evaluation across diverse scenarios, the 3CSim supports the learning and generalization of AD systems, contributing to their safety and reliability in real-world conditions. \par

\begin{figure*}[!ht]
	\centering
\includegraphics[width=1\textwidth]{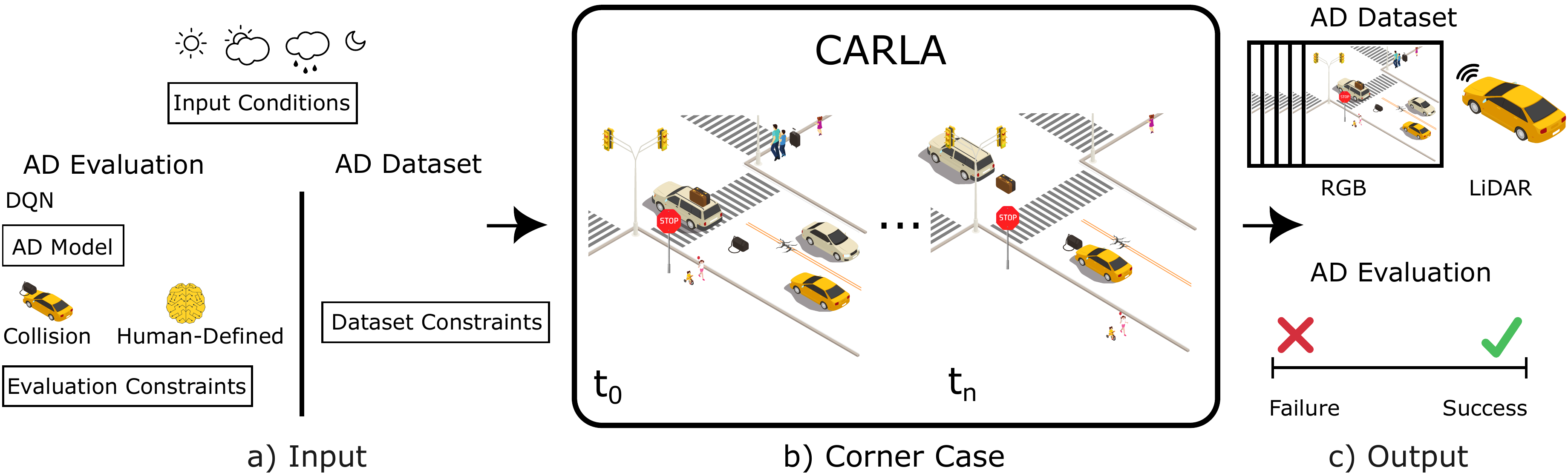}
    	\caption{Overview of the 3CSim framework for AD system evaluation, including a) input configuration, b) corner case-triggered simulation, and c) output as either scenario assessment or data for further analysis.}
	\label{fig:3csim}
\end{figure*}

Fig.~\ref{fig:3csim}~a) presents the inputs to our proposed simulation. The most fundamental are input conditions describing the urban environment and scenario parameters, such as vehicle speed, weather conditions, and traffic density. The 3CSim serves two primary functions: evaluating AD models and generating datasets of corner cases for subsequent analysis. For AD system evaluation, an AD model, such as a reinforcement learning (RL) model like DQN, is loaded as the key participant in the scenario. Evaluation constraints are then established to determine the success or failure of the scenario, either through simple collision detection or a hierarchical system where collisions are weighted differently, such as prioritizing collisions with humans as more severe than those with small traffic signs. To generate datasets for further analysis, the corner case is executed in the CARLA simulator, data is collected according to predefined dataset constraints, and then saved in specified formats. \par

Once all input parameters are defined, the simulation is executed from a predefined start time $t_0$ to a specified end time $t_n$. The simulation may conclude either after a set duration or upon the occurrence of specific events, such as a collision or the vehicle remaining stationary for a designated period. Fig.~\ref{fig:3csim}~b) illustrates a scenario where the AD model, represented by the yellow vehicle, must decide whether to go straight, turn right, turn left, or stop. In this instance, the simulation ended due to a collision with a piece of luggage that had fallen onto the road. The scenario can be repeated with identical conditions or modified by altering variables such as weather conditions or the timing of unexpected events, such as adjusting the moment when the luggage falls onto the road. \par

Fig.~\ref{fig:3csim}~c) illustrates the outputs of the 3CSim. For AD model evaluation, the output includes metrics that determine the success or failure of the model based on the predefined evaluation constrains from the input. For dataset generation, the output consists of data from the simulated urban environment, including sensor data such as RGB frames and LiDAR, which are compiled into datasets for subsequent processing using advanced AI techniques. \par
The code for our 3CSim is publicly available at a GitHub repository \footnote{\url{https://github.com/Maftej/3csim}}. \par

\section{Taxonomy of Corner Cases}
\label{sec:taxonomy}
\begin{figure*}[!htp]
	\centering
\includegraphics[width=1\textwidth]{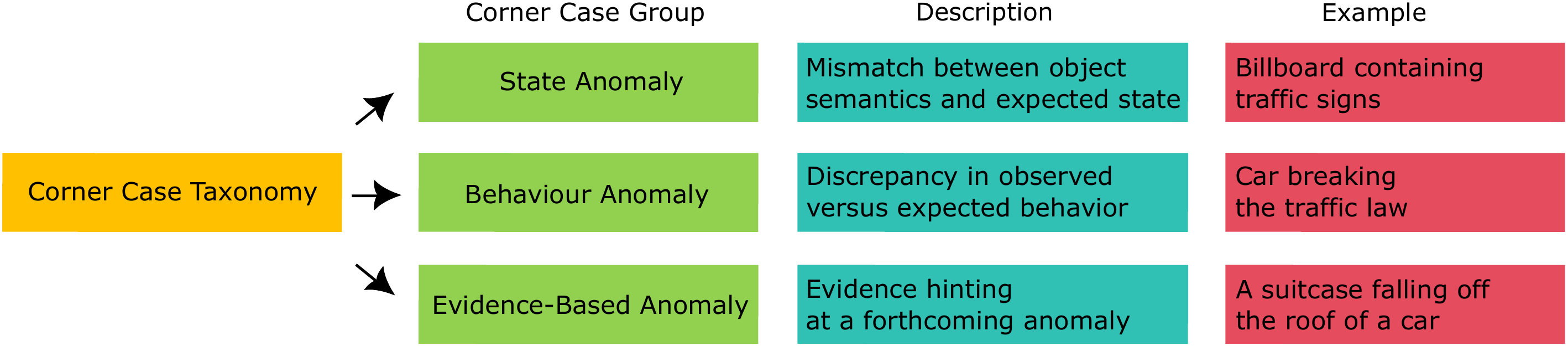}
    	\caption{A taxonomy of corner cases categorized into state anomalies, behavior anomalies, and evidence-based anomalies.}
	\label{fig:2c_taxonomy}
\end{figure*}

This section presents a taxonomy of corner cases, comprising 32 unique scenarios with variations defined by the simulator's boundaries, such as 9 predefined weather conditions, timing, or traffic density. Fig.~\ref{fig:2c_taxonomy} illustrates three types of corner cases included. Within this simulation, rare traffic events are classified as anomalies, typically associated with objects exhibiting unusual semantics or behavior. The final category, evidence-based anomalies, involves observable indicators that prompt caution, enabling proactive measures before the corner case occurs. In this subsection, each anomaly is described in detail, along with relevant examples.

\subsection{State Anomaly}
\label{state_anomaly}

\begin{figure}[!htp]
	\centering
\includegraphics[width=0.48\textwidth]{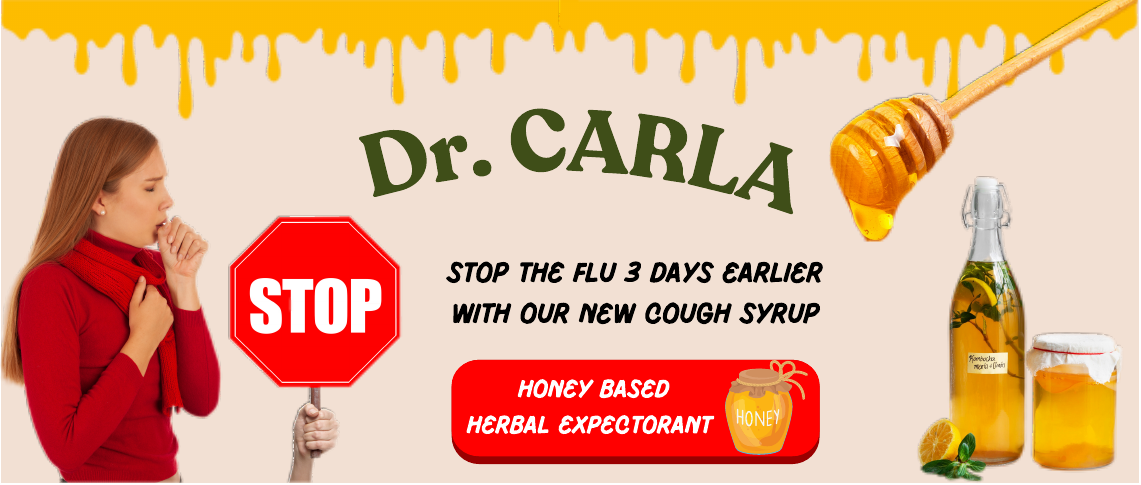}
    	\caption{A STOP sign integrated into an advertisement.}
	\label{fig:stop_sign_ad}
\end{figure}

State anomalies occur when there is a mismatch between the semantics of an object and its expected state, leading to potential misinterpretations by a system. Specifically, these anomalies arise when objects alter their appearance or context, falsely suggesting a different function than intended. A typical example is advertisement placed near roads that resemble traffic sign STOP as in Fig.~\ref{fig:stop_sign_ad}.

In our 3CSim, several corner cases are implemented to address these challenges. For instance, a video advertisement depicting a pedestrian holding a soft drink with the brand name "Carla Cola" prominently displayed could be misidentified as an actual pedestrian by a vehicle's detection system. Similarly, a billboard promoting a party that features an image of a traffic light may be mistaken for a real traffic signal, leading the vehicle to inappropriately stop or change speed.

Another case is a billboard displaying the word "stop" in the context of sneezing, which could be misinterpreted as a genuine STOP sign, causing the vehicle to halt unexpectedly. Likewise, a sign indicating "parking ahead" but specifying "only for bar members" might mislead the vehicle into considering it a general parking sign, creating confusion in locating proper parking.

\begin{figure}[!htp]
	\centering
\includegraphics[width=0.48\textwidth]{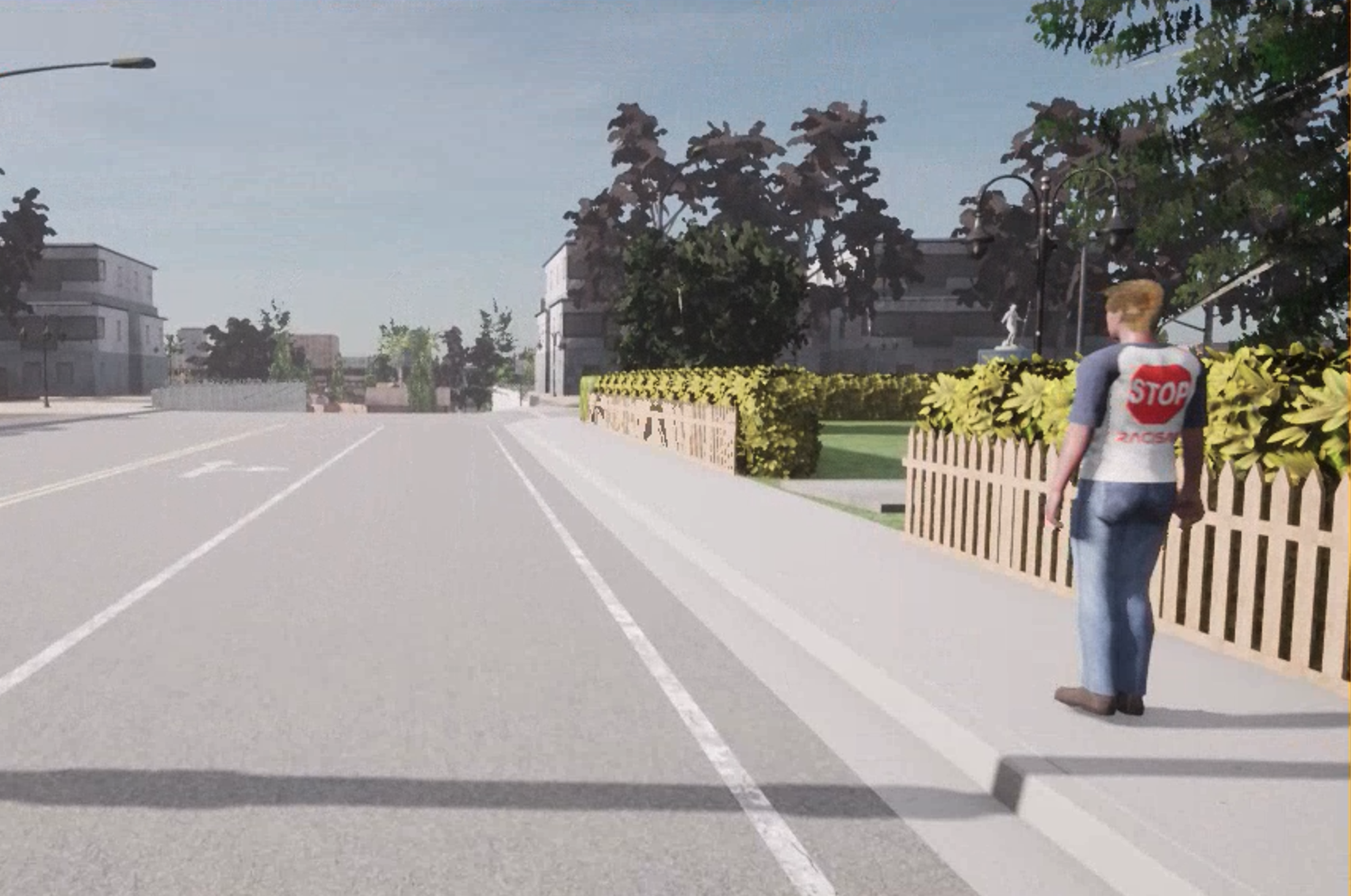}
    	\caption{A pedestrian wearing T-shirt displaying the traffic sign STOP.}
	\label{fig:stop_tshirt}
\end{figure}

Moreover, an advertisement showing a turn symbol for a soft drink could be mistaken for an actual turn sign, potentially resulting in unnecessary vehicle maneuvers. A billboard with the message "Yield - to Fun" could be interpreted as a real yield sign, causing the vehicle to yield inappropriately. Additionally, an advertisement featuring a green traffic light with the phrase "go for sale" might be confused with a real traffic signal, prompting the vehicle to proceed when it should not. A billboard stating "stop" for dinner could similarly lead to an unintended stop if misinterpreted as a genuine traffic STOP sign. Another example is a pedestrian wearing a T-shirt displaying the traffic sign, as shown in Fig.~\ref{fig:stop_tshirt}.

These examples illustrate how state anomalies, particularly in the context of road signs and advertisements, can pose significant challenges to autonomous systems, potentially compromising safety and functionality.

\subsection{Behavior Anomaly}
\label{behavior_anomaly}

Behavior anomalies refer to discrepancies between observed and expected behaviors, often leading to unpredictability and challenges in system response. These anomalies occur when objects deviate from typical behavior, such as a car breaking traffic laws, making the situation less predictable and more difficult for autonomous systems to handle.

For example, a car crash scenario obstructing a lane requires the vehicle's perception system to recognize the incident from a distance, calculate an alternate route, and safely change lanes if necessary. In the case of active emergency vehicles blocking an exit at a roundabout, the autonomous system must recognize the special status of these vehicles and reroute accordingly, while still adhering to road rules and vehicle prioritization.

Another complex situation involves a police car chase, where the AV must recognize the urgency and nature of the chase, possibly altering its route or pulling over to avoid interfering with law enforcement activities. Unusual pedestrian or animal behavior, such as a pedestrian standing near a crosswalk without crossing, can confuse prediction models that anticipate pedestrian movements based on proximity to crosswalks.

Erratic biker maneuvers beside the vehicle present a continuous threat, requiring the autonomous system to continuously update its path planning and speed to account for potential sudden turns by the biker. Similarly, a loose shopping cart rolling downhill into the vehicle's path represents a dynamic obstacle that the vehicle must detect and evade without prior warning, highlighting the need for real-time obstacle recognition and avoidance capabilities.

\begin{figure}[!htp]
	\centering
\includegraphics[width=0.48\textwidth]{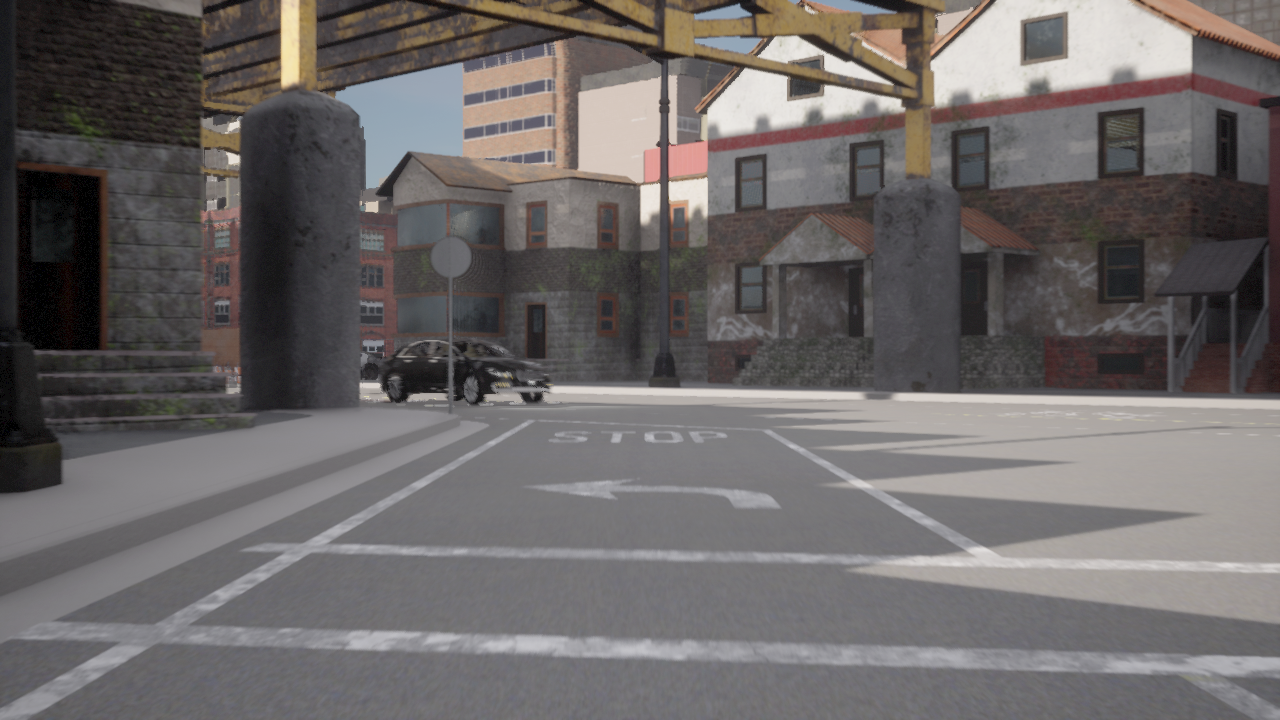}
    	\caption{An oncoming vehicle is indicating to turn right into one-way street.}
	\label{fig:oncoming_car_one_way}
\end{figure}

A particularly risky scenario is when a car drives against the vehicle on a one-way street as depicted in Fig.~\ref{fig:oncoming_car_one_way}. Here, the vehicle must anticipate the opposing vehicle's potential actions, such as stopping after realizing the mistake or turning around. Finally, a scenario where a ball is launched over an obstacle onto a high-speed road could cause an AV to brake suddenly to avoid a potential collision.

Each of these scenarios demands that the vehicle's perception algorithms interpret not just the current frame but also the context provided by a sequence of frames. This multi-frame analysis helps in understanding the progression of dynamic elements within the scene, enabling more accurate predictions and safer maneuvering decisions. In our dataset, we introduce these complex scenarios using CARLA's simulation environment, positioning emergency vehicles like firetrucks, ambulances, and police vehicles in non-standard settings to rigorously test the adaptability and accuracy of AD systems under varied, unpredictable conditions.

While most of these scenarios are not inherently dangerous, they require precise action to ensure safety. Evaluations can be conducted based on the vehicle's response to avoid collisions in these scenarios, providing insights into the system's effectiveness in handling behavior anomalies.

\subsection{Evidence-Based Anomaly}
\label{evidence_based_anomaly}

Evidence-based anomalies in AD refer to situations where there is prior evidence indicating a potential forthcoming anomaly, allowing the vehicle to prepare or adjust its strategy to mitigate hazards. These scenarios demand both scene and temporal understanding, requiring the autonomous system to process a sequence of images to detect and respond effectively to irregular or unexpected events. The presence of early indicators in these situations provides the opportunity for the vehicle to anticipate and manage the anomaly before it fully manifests.

\begin{figure}[!htp]
	\centering
\includegraphics[width=0.48\textwidth]{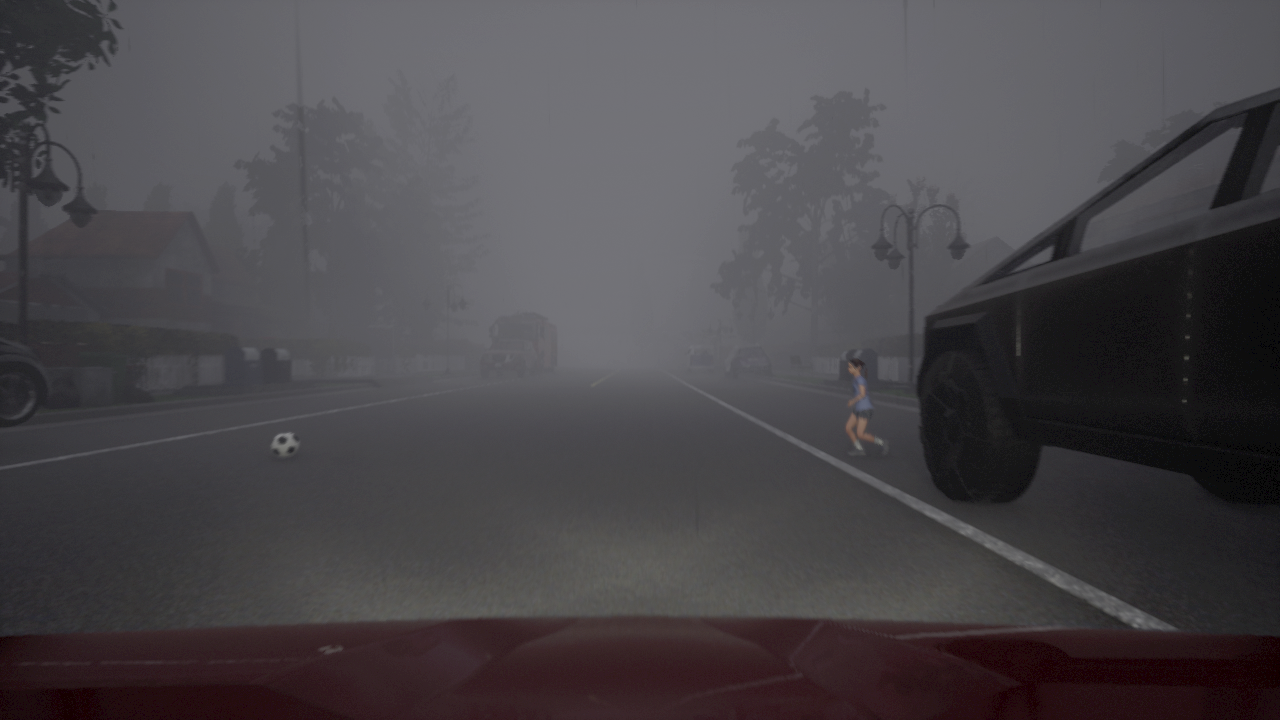}
    	\caption{A soccer ball as an indicator of a child's imminent entry onto the road.}
	\label{fig:ball_evidence}
\end{figure}

\begin{figure}[!htp]
	\centering
\includegraphics[width=0.48\textwidth]{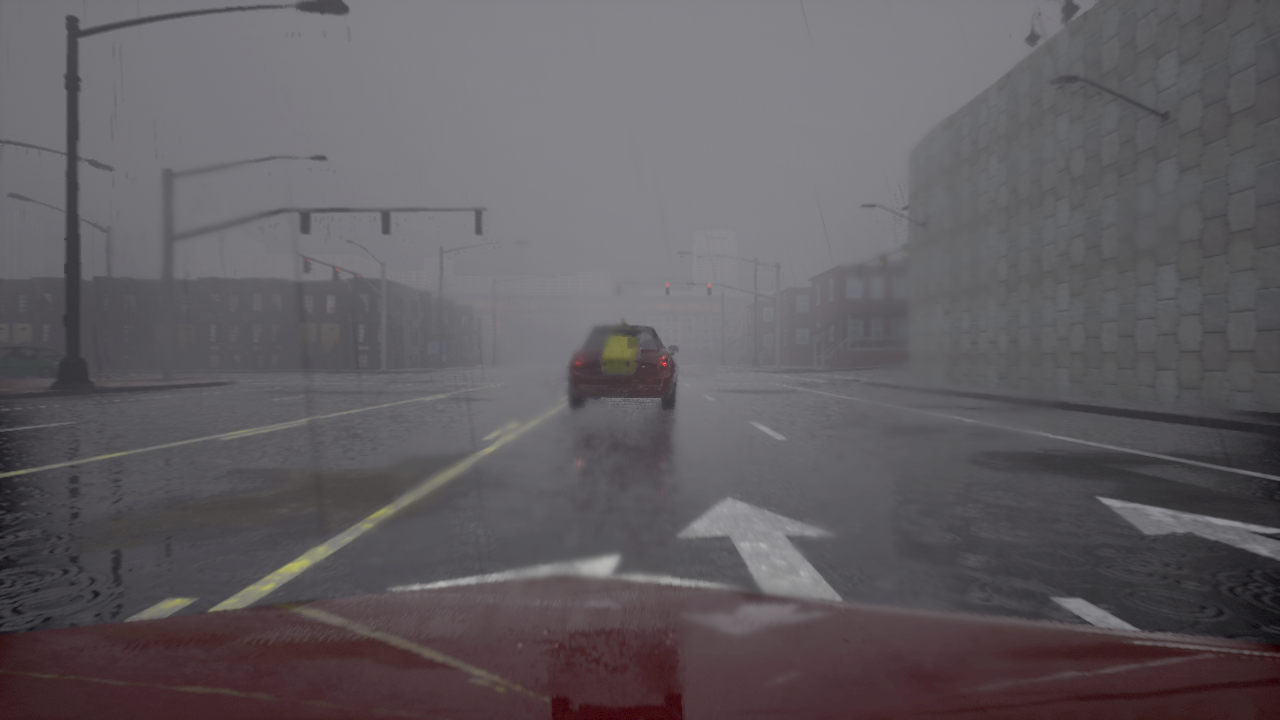}
    	\caption{A luggage may fall onto the road causing sudden appearance of foreign object.}
	\label{fig:luggage_fall}
\end{figure}

As depicted in Fig.~\ref{fig:ball_evidence}, parked cars obscure the vehicle’s vision and a child suddenly runs into the street chasing a football. There is a dual hazard: the blocked line of sight and the unexpected appearance of the child. The AV must rapidly assess the situation and take appropriate action to avoid a potentially fatal accident. Similarly, Fig.~\ref{fig:luggage_fall} depicts luggage falling from the roof of a car, the vehicle must quickly decide whether to swerve, brake, or continue, depending on the object’s trajectory and the road conditions.

In another scenario, if a car parks too close to the vehicle's lane and a door opens suddenly, the vehicle must predict this behavior and react swiftly to avoid a side collision. A similar challenge arises when a worker runs out from behind a parked van, requiring immediate recognition and response to prevent an accident, highlighting the importance of peripheral detection in cluttered environments.

Moreover, if a courier causes a barrel to fall from a hand truck into the vehicle’s lane, the system must make a quick decision to either swerve or brake to avoid a collision. An EMS vehicle with lights flashing, departing from a hospital while disregarding traffic lights and speeding, presents a risky scenario where the AV must yield and maneuver safely to avoid an accident. 

These evidence-based anomalies demonstrate the critical need for advanced perception systems in AVs that can interpret complex situations and respond appropriately. The early detection of potential hazards allows the vehicle to take preemptive actions to maintain safety, not only for its passengers but also for surrounding traffic and pedestrians.

\section{Conclusion}
\label{conclusion}
In this work, we propose the 3CSim
for control assessment in AD. This simulation is designed for evaluating AD systems within the CARLA simulator. The scenarios can be repeated under identical conditions or with slight modifications, enabling unique assessments that are not feasible in real-world environments. Data from these simulations are collected to create a dataset for further processing. Additionally, we introduce a taxonomy of corner cases, categorized into three groups: state anomalies, behavior anomalies, and evidence-based anomalies. We implemented 32 unique corner cases, with modifiable parameters such as 9 predefined weather conditions, timing, and traffic density. Future work will focus on extending these corner cases, enhancing simulation flexibility by incorporating more input parameters, and developing custom evaluation metrics for corner cases.

\section*{Acknowledgment}
\label{acknowledgment}
This work was supported by the Ministry of Education, Science, Research and Sport of the Slovak Republic, and the Slovak Academy of Sciences under Grant VEGA 1/0685/23 and by the Slovak Research and Development Agency under Grant APVV SK-CZ-RD-21-0028 and APVV-23-0512, and by Research and Innovation Authority VAIA under Grant 09I03-03-V04-00395.

\bibliography{ref.bib}{}

\begin{thebibliography}{10}
\providecommand{\url}[1]{#1}
\csname url@samestyle\endcsname
\providecommand{\newblock}{\relax}
\providecommand{\bibinfo}[2]{#2}
\providecommand{\BIBentrySTDinterwordspacing}{\spaceskip=0pt\relax}
\providecommand{\BIBentryALTinterwordstretchfactor}{4}
\providecommand{\BIBentryALTinterwordspacing}{\spaceskip=\fontdimen2\font plus
\BIBentryALTinterwordstretchfactor\fontdimen3\font minus \fontdimen4\font\relax}
\providecommand{\BIBforeignlanguage}[2]{{%
\expandafter\ifx\csname l@#1\endcsname\relax
\typeout{** WARNING: IEEEtran.bst: No hyphenation pattern has been}%
\typeout{** loaded for the language `#1'. Using the pattern for}%
\typeout{** the default language instead.}%
\else
\language=\csname l@#1\endcsname
\fi
#2}}
\providecommand{\BIBdecl}{\relax}
\BIBdecl

\bibitem{chib2024recent}
P.~S. Chib and P.~Singh, ``Recent advancements in end-to-end autonomous driving using deep learning: A survey,'' \emph{IEEE Transactions on Intelligent Vehicles}, vol.~9, no.~1, pp. 103--118, 2024.

\bibitem{luo2022jfp}
\BIBentryALTinterwordspacing
W.~Luo, C.~Park, A.~Cornman, B.~Sapp, and D.~Anguelov, ``{JFP}: Joint future prediction with interactive multi-agent modeling for autonomous driving,'' in \emph{6th Annual Conference on Robot Learning}, 2022. [Online]. Available: \url{https://openreview.net/forum?id=Y42uoIekm5b}
\BIBentrySTDinterwordspacing

\bibitem{siu2023motion}
S.~Teng, X.~Hu, P.~Deng, B.~Li, Y.~Li, Y.~Ai, D.~Yang, L.~Li, Z.~Xuanyuan, F.~Zhu, and L.~Chen, ``Motion planning for autonomous driving: The state of the art and future perspectives,'' \emph{IEEE Transactions on Intelligent Vehicles}, vol.~8, no.~6, pp. 3692--3711, 2023.

\bibitem{ravi2022deep}
B.~R. Kiran, I.~Sobh, V.~Talpaert, P.~Mannion, A.~A.~A. Sallab, S.~Yogamani, and P.~Pérez, ``Deep reinforcement learning for autonomous driving: A survey,'' \emph{IEEE Transactions on Intelligent Transportation Systems}, vol.~23, no.~6, pp. 4909--4926, 2022.

\bibitem{jingwei2023eventtriggered}
J.~Lu, L.~Han, Q.~Wei, X.~Wang, X.~Dai, and F.-Y. Wang, ``Event-triggered deep reinforcement learning using parallel control: A case study in autonomous driving,'' \emph{IEEE Transactions on Intelligent Vehicles}, vol.~8, no.~4, pp. 2821--2831, 2023.

\bibitem{jianyu2019deep}
J.~Chen, B.~Yuan, and M.~Tomizuka, ``Deep imitation learning for autonomous driving in generic urban scenarios with enhanced safety,'' in \emph{2019 IEEE/RSJ International Conference on Intelligent Robots and Systems (IROS)}, 2019, pp. 2884--2890.

\bibitem{siyu2023hierarchical}
S.~Teng, L.~Chen, Y.~Ai, Y.~Zhou, Z.~Xuanyuan, and X.~Hu, ``Hierarchical interpretable imitation learning for end-to-end autonomous driving,'' \emph{IEEE Transactions on Intelligent Vehicles}, vol.~8, no.~1, pp. 673--683, 2023.

\bibitem{bogdoll2022anomaly}
D.~Bogdoll, M.~Nitsche, and J.~M. Zöllner, ``Anomaly detection in autonomous driving: A survey,'' in \emph{2022 IEEE/CVF Conference on Computer Vision and Pattern Recognition Workshops (CVPRW)}, 2022, pp. 4487--4498.

\bibitem{hanh2023adslead}
X.~Han, Y.~Zhou, K.~Chen, H.~Qiu, M.~Qiu, Y.~Liu, and T.~Zhang, ``Ads-lead: Lifelong anomaly detection in autonomous driving systems,'' \emph{IEEE Transactions on Intelligent Transportation Systems}, vol.~24, no.~1, pp. 1039--1051, 2023.

\bibitem{bogdoll2021description}
D.~Bogdoll, J.~Breitenstein, F.~Heidecker, M.~Bieshaar, B.~Sick, T.~Fingscheidt, and J.~M. Zöllner, ``Description of corner cases in automated driving: Goals and challenges,'' in \emph{2021 IEEE/CVF International Conference on Computer Vision Workshops (ICCVW)}, 2021, pp. 1023--1028.

\bibitem{fu2024drive}
D.~Fu, X.~Li, L.~Wen, M.~Dou, P.~Cai, B.~Shi, and Y.~Qiao, ``Drive like a human: Rethinking autonomous driving with large language models,'' in \emph{2024 IEEE/CVF Winter Conference on Applications of Computer Vision Workshops (WACVW)}, 2024, pp. 910--919.

\bibitem{hossain2023autonomous}
\BIBentryALTinterwordspacing
J.~Hossain, ``Autonomous driving with deep reinforcement learning in carla simulation,'' 2023. [Online]. Available: \url{https://arxiv.org/abs/2306.11217}
\BIBentrySTDinterwordspacing

\bibitem{nehme2023safe}
\BIBentryALTinterwordspacing
G.~Nehme and T.~Y. Deo, ``Safe navigation: Training autonomous vehicles using deep reinforcement learning in carla,'' 2023. [Online]. Available: \url{https://arxiv.org/abs/2311.10735}
\BIBentrySTDinterwordspacing

\bibitem{kashyap2023transfuser}
K.~Chitta, A.~Prakash, B.~Jaeger, Z.~Yu, K.~Renz, and A.~Geiger, ``Transfuser: Imitation with transformer-based sensor fusion for autonomous driving,'' \emph{IEEE Transactions on Pattern Analysis and Machine Intelligence}, vol.~45, no.~11, pp. 12\,878--12\,895, 2023.

\bibitem{niu2023stackelberg}
\BIBentryALTinterwordspacing
H.~Niu, Q.~Chen, Y.~Li, Y.~Zhang, and J.~Hu, ``Stackelberg driver model for continual policy improvement in scenario-based closed-loop autonomous driving,'' 2023. [Online]. Available: \url{https://arxiv.org/abs/2309.14235}
\BIBentrySTDinterwordspacing

\bibitem{drayson2024ccsgg}
\BIBentryALTinterwordspacing
G.~Drayson, E.~Panagiotaki, D.~Omeiza, and L.~Kunze, ``Cc-sgg: Corner case scenario generation using learned scene graphs,'' 2024. [Online]. Available: \url{https://arxiv.org/abs/2309.09844}
\BIBentrySTDinterwordspacing

\bibitem{daoud2024cornersim}
\BIBentryALTinterwordspacing
A.~Daoud, C.~Bunel, and M.~Guériau, ``Cornersim: A virtualization framework to generate realistic corner-case scenarios for autonomous driving perception testing,'' \emph{Procedia Computer Science}, vol. 238, pp. 184--191, 2024, the 15th International Conference on Ambient Systems, Networks and Technologies Networks (ANT) / The 7th International Conference on Emerging Data and Industry 4.0 (EDI40), April 23-25, 2024, Hasselt University, Belgium. [Online]. Available: \url{https://www.sciencedirect.com/science/article/pii/S187705092401250X}
\BIBentrySTDinterwordspacing

\bibitem{li2024first}
\BIBentryALTinterwordspacing
L.~Li, H.~Wu, B.~Yao, T.~He, S.~Huang, and C.~Liu, ``First-principles based 3d virtual simulation testing for discovering sotif corner cases of autonomous driving,'' 2024. [Online]. Available: \url{https://arxiv.org/abs/2401.11876}
\BIBentrySTDinterwordspacing

\bibitem{zhou2023corner}
J.~Zhou and J.~Beyerer, ``Corner cases in data-driven automated driving: Definitions, properties and solutions,'' in \emph{2023 IEEE Intelligent Vehicles Symposium (IV)}, 2023, pp. 1--8.

\bibitem{breitenstein2020systematization}
J.~Breitenstein, J.-A. Termöhlen, D.~Lipinski, and T.~Fingscheidt, ``Systematization of corner cases for visual perception in automated driving,'' in \emph{2020 IEEE Intelligent Vehicles Symposium (IV)}, 2020, pp. 1257--1264.

\bibitem{heidecker2021applicationdriven}
\BIBentryALTinterwordspacing
F.~Heidecker, J.~Breitenstein, K.~Rosch, J.~Lohdefink, M.~Bieshaar, C.~Stiller, T.~Fingscheidt, and B.~Sick, ``An application-driven conceptualization of corner cases for perception in highly automated driving,'' in \emph{2021 IEEE Intelligent Vehicles Symposium (IV)}.\hskip 1em plus 0.5em minus 0.4em\relax IEEE, Jul. 2021. [Online]. Available: \url{http://dx.doi.org/10.1109/IV48863.2021.9575933}
\BIBentrySTDinterwordspacing

\bibitem{pfeil2022on}
J.~Pfeil, J.~Wieland, T.~Michalke, and A.~Theissler, ``On why the system makes the corner case: Ai-based holistic anomaly detection for autonomous driving,'' in \emph{2022 IEEE Intelligent Vehicles Symposium (IV)}, 2022, pp. 337--344.

\end{thebibliography}
\bibliographystyle{IEEEtran}

\end{document}